\pdfoutput=1

\documentclass[11pt]{article}

\usepackage[final]{acl}

\usepackage{times}
\usepackage{latexsym}
\usepackage{hyperref}
\usepackage{url}
\usepackage{colortbl}  
\usepackage{xcolor}
\usepackage{latexsym}
\usepackage{multirow}
\usepackage{subfigure}
\usepackage{amsmath}
\usepackage{booktabs}
\usepackage{amssymb}
\usepackage{microtype}
\usepackage{balance}
\usepackage{inconsolata}
\usepackage{adjustbox}
\usepackage{graphicx}

\usepackage[utf8]{inputenc} 
\usepackage[T1]{fontenc}    
\usepackage{hyperref}       
\usepackage{url}            
\usepackage{booktabs}       
\usepackage{amsfonts}       
\usepackage{nicefrac}       
\usepackage{microtype}      
\usepackage{xcolor}         
\usepackage{multirow}
\usepackage{subfigure}
\usepackage[english]{babel}
\usepackage{moresize}
\usepackage{amsmath}
\usepackage{algorithmic}
\usepackage{balance}
\usepackage{comment}
\usepackage{paralist}
\usepackage{bm}
\usepackage{pgfplots}
\usepackage{flushend}
\usepackage[english]{babel}
\usepackage{graphicx}

\usepackage{amssymb}
\usepackage{amsfonts}
\usepackage{url}
\usepackage{bbm}
\usepackage{hhline}
\usepackage{longtable}
\usepackage{rotating}
\usepackage{multirow}
\usepackage{mathrsfs}
\usepackage{enumitem}
\usepackage[linesnumbered,algoruled,boxed,lined]{algorithm2e}
\usepackage{adjustbox}
\usepackage{hyperref}
\usepackage{pgfplots}
\usepackage{wrapfig}
\usepackage{caption}
\usepackage[T1]{fontenc}

\usepackage[utf8]{inputenc}

\usepackage{microtype}

\usepackage{inconsolata}

\usepackage[linesnumbered,algoruled,boxed,lined]{algorithm2e}
\usepackage{listings}

\usepackage[most]{tcolorbox}
\tcbset{aibox/.style={
    breakable,
    break at=\maxdimen,
    fontlower=\scriptsize, 
}}
\newtcolorbox{AIbox}[2][]{aibox,title={#2},#1}
\definecolor{lightgrey}{HTML}{E7E7E7}

\title{Retrieval-Augmented Generation with Hierarchical Knowledge}

\author{
  Haoyu Huang$^{1,2}$\footnotemark[1], Yongfeng Huang$^{2}$\footnotemark[1], Junjie Yang$^{2}$, Zhenyu Pan$^{1,2}$, Yongqiang Chen$^{1}$\\
  \textbf{Kaili Ma$^{1}$, Hongzhi Chen$^{1}$, James Cheng$^{2}$}  \\
  $^1$KASMA.ai\\
  $^2$CSE, The Chinese University of Hong Kong \\
  \texttt{\{haoyuhuang,zhenyupan,yqchen,klma,chenhongzhi\}@kasma.ai} \\
  \texttt{\{haoyuhuang,zhenyupan,1155215805\}@link.cuhk.edu.hk} \\
  \texttt{\{yfhuang22,jcheng\}@cse.cuhk.edu.hk} \\
}

\begin{document}
\maketitle
{
  \renewcommand{\thefootnote}%
  {\fnsymbol{footnote}}
  \footnotetext[1]{Equal contribution. This research was conducted at kasma.ai.}
}
\begin{abstract}
Graph-based Retrieval-Augmented Generation (RAG) methods have significantly enhanced the performance of large language models (LLMs) in domain-specific tasks. However, existing RAG methods do not adequately utilize the naturally inherent hierarchical knowledge in human cognition, which limits the capabilities of RAG systems. In this paper, we introduce a new RAG approach, called HiRAG, which  utilizes hierarchical knowledge to enhance the semantic understanding and structure capturing capabilities of RAG systems in the indexing and retrieval processes. 
Our extensive experiments demonstrate that HiRAG achieves significant performance improvements over the state-of-the-art baseline methods. The code of our proposed method is available at \href{https://github.com/hhy-huang/HiRAG}{https://github.com/hhy-huang/HiRAG}.

\end{abstract}

\section{Introduction}\label{sec:intro}

\begin{figure*}[htbp]
\centerline{\includegraphics[width=1\linewidth]{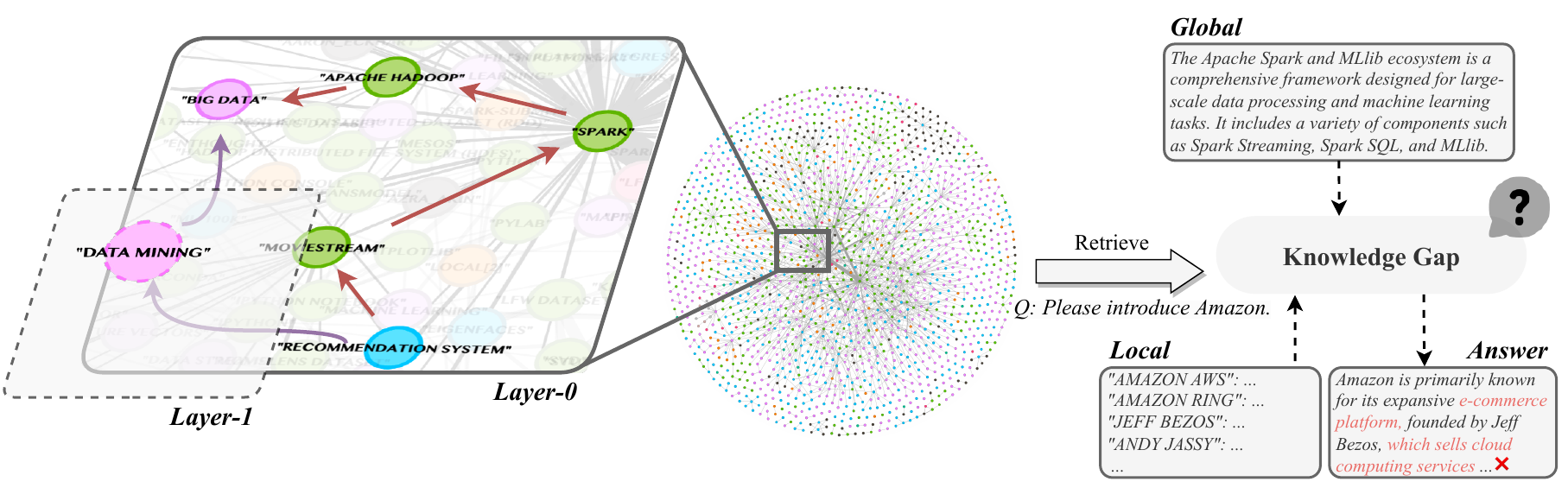}}
\caption{The challenges faced by existing RAG systems: (1) Distant structural relationship between semantically similar entities. (2) Knowledge gap between  local and global knowledge.}
\label{fc}
\end{figure*}



Retrieval Augmented Generation (RAG)~\cite{gao2023retrieval, lewis2020retrieval, fan2024survey}  has been introduced to enhance the capabilities of LLMs in domain-specific or knowledge-intensive tasks. Naive RAG methods retrieve text chunks that are relevant to a query, which serve as references for LLMs to generate responses, thus helping address the problem of "Hallucination"~\cite{zhang2023siren, tang2024multihoprag}. 
However, naive RAG methods usually overlook the relationships among entities in the retrieved text chunks. 
To address this issue, RAG systems with graph structures were proposed~\cite{edge2024local, liang2024kagboostingllmsprofessional, zhang2025surveygraphretrievalaugmentedgeneration, peng2024graphretrievalaugmentedgenerationsurvey}, which construct knowledge graphs (KGs) to model relationships between entities in the input documents. 
Although existing RAG systems integrating graph structures have demonstrated outstanding performance on various tasks, they still have some serious limitations. GraphRAG~\cite{edge2024local} introduces communities in indexing using the Leiden algorithm~\cite{traag2019louvain}, but the communities only capture the structural proximity of the entities in the KG.  KAG~\cite{liang2024kagboostingllmsprofessional} indexes with a hierarchical representation of information and knowledge, but their hierarchical structure relies too much on manual annotation and requires a lot of human domain knowledge, which renders their method not scalable to general tasks. LightRAG~\cite{guo2024lightrag} utilizes a dual-level retrieval approach to obtain local and global knowledge as the contexts for a query, but it ignores the \textbf{knowledge gap} between local and global knowledge, that is, 
local knowledge represented by the retrieved individual entities (i.e., entity-specific details) may not be semantically related to the global knowledge represented in the retrieved community summaries (i.e.,  broader, aggregated summaries), as these individual entities may not be a part of the retrieved communities for a query.

We highlight two critical challenges in existing RAG systems that integrate graph structures: \textbf{(1)~distant structural relationship between semantically similar entities} and \textbf{(2)~knowledge gap between local and global knowledge}.  We illustrate them using a real example from a public dataset, as shown in Figure~\ref{fc}. 

Challenge (1) occurs because existing methods over-rely on source documents, often resulting in constructing a knowledge graph (KG) with many entities that are not structurally proximate in the KG even though they share semantically similar attributes. For example, in Figure~\ref{fc}, although the entities "BIG DATA" and "RECOMMENDATION SYSTEM" share semantic relevance under the concept of "DATA MINING", their distant structural relationship in the KG reflects a corpus-driven disconnect. 
These inconsistencies between semantic relevance and structural proximity are systemic in KGs, undermining their utility in RAG systems where contextual coherence is critical. 

Challenge (2) occurs as existing methods~\cite{guo2024lightrag, edge2024local} typically retrieve context either from global or local perspectives but fail to address the inherent disparity between these knowledge layers. Consider the query "Please introduce Amazon" in Figure~\ref{fc}, where global context emphasizes Amazon's involvement in technological domains like big data and cloud computing, but local context retrieves entities directly linked to Amazon (e.g., subsidiaries, leadership). When these two knowledge layers are fed into LLMs as the contexts of a query without contextual alignment, LLMs may struggle to reconcile their distinct scopes, leading to disjointed reasoning, incomplete answers, or even contradictory outputs. For instance, an LLM might conflate Amazon’s role as a cloud provider (global) with its e-commerce operations (local), resulting in incoherent or factually inconsistent responses as the red words shown in the case. This underscores the need for new methods that bridge hierarchical knowledge layers to ensure cohesive reasoning in RAG systems.

To address these challenges, we propose \textbf{Retrieval-Augmented Generation with Hierarchical Knowledge (HiRAG)}, which integrates hierarchical knowledge into the indexing and retrieval processes. Hierarchical knowledge~\cite{sarrafzadeh2017improving} is a natural concept in both graph structure and human cognition, yet it has been overlooked in existing approaches. 
Specifically, to address Challenge~(1), we introduce \textbf{Indexing with Hierarchical Knowledge (HiIndex)}. Rather than simply constructing a flat KG, we index a KG hierarchically layer by layer. 
Each entity (or node) in a higher layer summarizes a cluster of entities in the lower layer, which can enhance the connectivity between semantically similar entities. For example, in Figure~\ref{fc}, the inclusion of the summary entity "DATA MINING" strengthens the connection between "BIG DATA" and "RECOMMENDATION SYSTEM". 
To address Challenge~(2), we propose \textbf{Retrieval with Hierarchical Knowledge (HiRetrieval)}. HiRetrieval effectively bridges local knowledge of entity descriptions to global knowledge of communities, thus resolving knowledge layer disparities. It provides a three-level context comprising the global level, the bridge level, and the local level knowledge to an LLM, enabling the LLM to  generate more comprehensive and precise responses. 

In summary, we make the following main contributions:
\begin{itemize}
    \item We identify and address two critical challenges in graph-based RAG systems: distant
    structural relationships between semantically similar entities and the knowledge gap between
    local and global information.
    \item We propose HiRAG, which introduces unsupervised hierarchical indexing and a novel bridging
    mechanism for effective knowledge integration, significantly advancing the state-of-the-art in
    RAG systems.
    \item Extensive experiments demonstrate both the effectiveness and efficiency of our approach,
    with comprehensive ablation studies validating the contribution of each component.
\end{itemize}
\section{Related Work }
In this section, we discuss recent research concerning graph-augmented LLMs, specifically RAG methods with graph structures. GNN-RAG~\cite{mavromatis2024gnn} employs GNN-based reasoning to retrieve query-related entities. Then they find the shortest path between the retrieved entities and candidate answer entities to construct reasoning paths. LightRAG~\cite{guo2024lightrag} integrates a dual-level retrieval method with graph-enhanced text indexing. They also decrease the computational costs and speed up the adjustment process. GRAG~\cite{hu2024grag} leverages a soft pruning approach to minimize the influence of irrelevant entities in retrieved subgraphs. It also implements prompt tuning to help LLMs comprehend textual and topological information in subgraphs by incorporating graph soft prompts. StructRAG~\cite{li2024structrag} identifies the most suitable structure for each task, transforms the initial documents into this organized structure, and subsequently generates responses according to the established structure. Microsoft GraphRAG~\cite{edge2024local} first retrieves related communities and then let the LLM generate the response with the retrieved communities. They also answer a query with global search and local search. KAG~\cite{liang2024kagboostingllmsprofessional} introduces a professional domain knowledge service framework and employs knowledge alignment using conceptual semantic reasoning to mitigate the noise issue in OpenIE. KAG also constructs domain expert knowledge using human-annotated schemas. ReG~\cite{zou2025weak} uses LLMs to refine weak supervision signals to align retrievers in fine-grained graph-based RAG systems.

\section{Preliminary and Definitions}

In this section, we give a general formulation of an RAG system with graph structure referring to the definitions in ~\citet{guo2024lightrag} and ~\citet{peng2024graph}.

In an RAG framework $\mathcal{M}$  as shown in Equation~\ref{eq:1}, $\text{LLM}$ is the generation module, $\mathcal{R}$ represents the retrieval module, $\varphi$ denotes the graph indexer, and $\psi$ refers to the graph retriever:
\begin{equation}
\mathcal{M} = (\text{LLM}, \mathcal{R}(\varphi, \psi)).
    \label{eq:1}
\end{equation}
When we answer a query, the answer we get from an RAG system is represented by $a^{*}$, which can be formulated as
\begin{equation}
a^{*} = arg\max_{a \in A}\mathcal{M}(a | q, \mathcal{G}),
    \label{eq:2}
\end{equation}
\begin{equation}
\mathcal{G} = \varphi(\mathcal{D}) = \{(h,r,t)|h, t \in \mathcal{V}, r \in \mathcal{E}\},
    \label{eq:3}
\end{equation}
where $\mathcal{M}(a | q, \mathcal{G})$ is the target distribution with a graph retriever $\psi(G | q, \mathcal{G})$ and a generator $\text{LLM}(a | q, G)$, and $A$ is the set of possible responses. The graph database $\mathcal{G}$ is constructed from the original external database $\mathcal{D}$. We utilize the total probability formula to decompose $\mathcal{M}(a|q,\mathcal{G})$, which can be expressed as
\begin{equation}
\mathcal{M}(a |q, \mathcal{G}) = \sum_{G \in \mathcal{G}} \text{LLM}(a|q,G)\cdot\psi(G|q,\mathcal{G}).
    \label{eq:4}
\end{equation}
Most of the time, we only need to retrieve the most relevant subgraph $G$ from the external graph database $\mathcal{G}$. Therefore, here we can approximate  $\mathcal{M}(a|q,\mathcal{G})$ as follows:
\begin{equation}
\mathcal{M}(a |q, \mathcal{G}) \approx \text{LLM}(a|q,G^{*})\cdot \psi(G^{*}|q,\mathcal{G}),
    \label{eq:5}
\end{equation}
where $G^{*}$ denotes the optimal subgraph we retrieve from the external graph database $\mathcal{G}$. What we finally want is to get a better generated answer $a^{*}$.
\begin{figure*}[htbp]
\centerline{\includegraphics[width=1\linewidth]{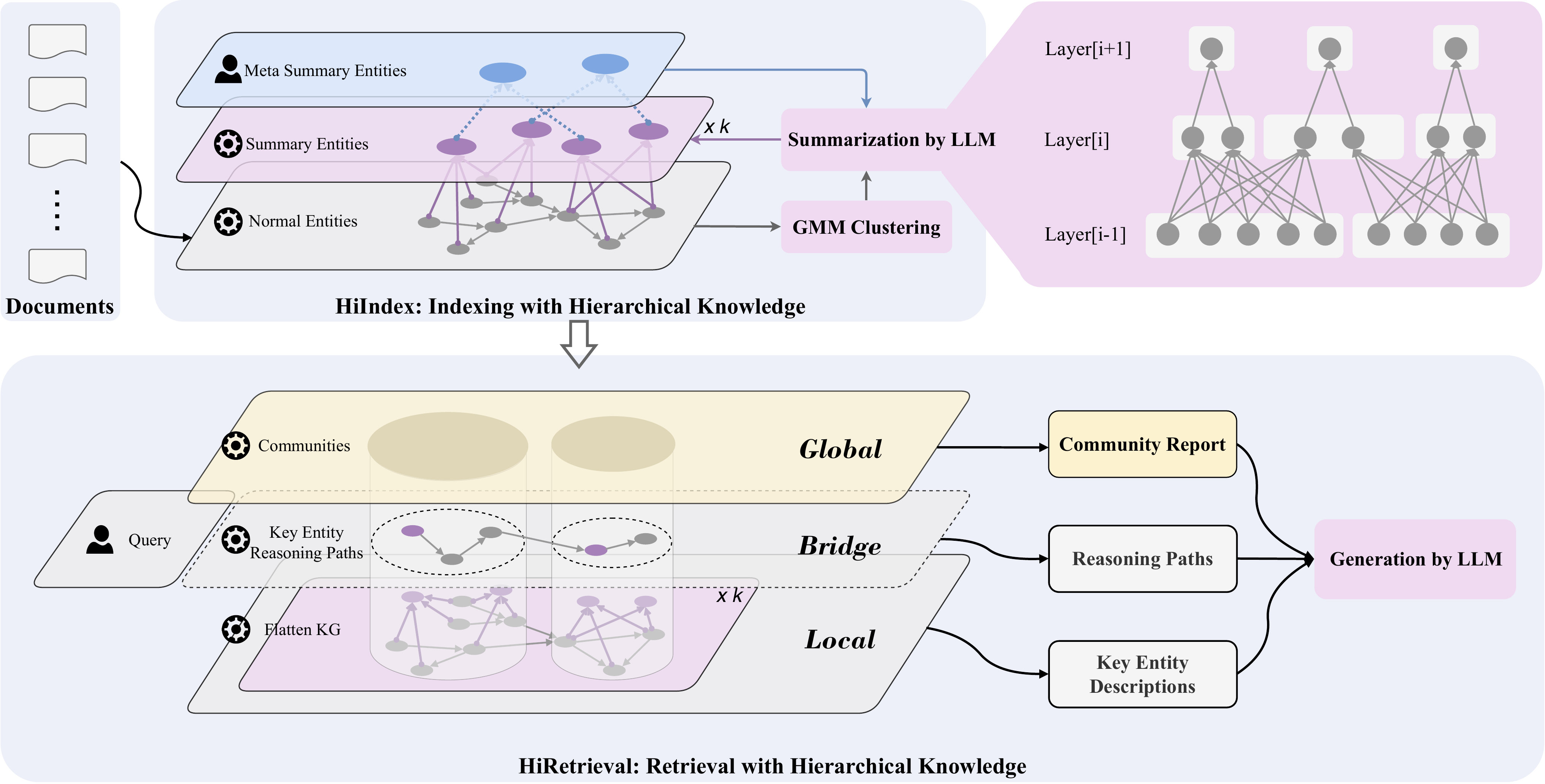}}
\caption{The overall architecture of the HiRAG framework.}
\label{fig:hirag_architecture}
\end{figure*}
\section{The HiRAG Framework}

HiRAG consists of the two modules, HiIndex and HiRetrieval, as shown in Figure~\ref{fig:hirag_architecture}. In the HiIndex module, we construct a hierarchical KG with different knowledge granularity in different layers. The summary entities in a higher layer represent more coarse-grained, high-level knowledge but they can enhance the connectivity between semantically similar entities in a lower layer. In the HiRetrieval module, we select the most relevant entities from each retrieved community and find the shortest path to connect them, which serve as the bridge-level knowledge to connect the knowledge at both local and global levels. Then an LLM will generate responses with these three-level knowledge as the context. 

\subsection{Indexing with Hierarchical Knowledge}\label{sec:hiindex}

In the HiIndex module, we index the input documents as a hierarchical KG. First, we employ the entity-centric triple extraction to construct a basic KG $\mathcal{G}_{0}$ following~\cite{carta2023iterative}. Specifically, we split the input documents into text chunks with some overlaps. These chunks will be fed into the LLM with well-designed prompts to extract entities $\mathcal{V}_{0}$ first. Then the LLM will generate relations (or edges) $\mathcal{E}_{0}$ between pairs of the extracted entities based on the information of the corresponding text chunks. The basic KG can be represented as
\begin{equation}
\mathcal{G}_{0} = \{(h,r,t) | h, t \in \mathcal{V}_{0}, r \in \mathcal{E}_{0}\}.
    \label{eq:6}
\end{equation}

The basic KG is also the 0-th layer of our hierarchical KG. We denote the set of entities (nodes) in the $i$-th layer as $\mathcal{L}_{i}$ where $\mathcal{L}_{0}=\mathcal{V}_{0}$. To construct the $i$-th layer of the hierarchical KG, for $i \ge 1$, we first fetch the embeddings of the entities in the $(i-1)$-th layer of the hierarchical KG, which is denoted as
\begin{equation}
\mathcal{Z}_{i-1} = \{\text{Embedding}(v) | v \in \mathcal{L}_{i-1}\},
    \label{eq:7}
\end{equation}

where $\text{Embedding}(v)$ is the embedding of an entity $v$. Then we employ Gaussian Mixture Models (GMMs) to conduct semantical clustering on $\mathcal{L}_{i-1}$ based on $\mathcal{Z}_{i-1}$, following the method described in RAPTOR~\cite{sarthi2024raptor}. We obtain a set of clusters as
\begin{equation}
\mathcal{C}_{i-1} = \text{GMM}(\mathcal{L}_{i-1}, \mathcal{Z}_{i-1}) = \{\mathcal{S}_{1}, \dots, \mathcal{S}_{c}\},
\label{eq:8}
\end{equation}
where $\forall x,y \in [1,c]$, $|\mathcal{S}_{x} \cap \mathcal{S}_{y}| \ge 0$ and $\bigcup_{1\le x \le c} \mathcal{S}_{x} = \mathcal{L}_{i-1}$. After clustering with GMMs, the descriptions of the entities in each cluster in $\mathcal{C}_{i-1}$ are fed into the LLM to generate a set of summary entities for the $i$-th layer. Thus, the set of summary entities in the $i$-th layer, i.e., $\mathcal{L}_{i}$, is the union of the sets of summary entities generated from all clusters in $\mathcal{C}_{i-1}$. Then, we create the relations between entities in $\mathcal{L}_{i-1}$ and entities in $\mathcal{L}_{i}$,  denoted as $\mathcal{E}_{\{i-1, i\}}$, by connecting the entities in each cluster $\mathcal{S} \in \mathcal{C}_{i-1}$ to the corresponding summary entities  in $\mathcal{L}_{i}$ that are generated from the entities in $\mathcal{S}$. 

To generate summary entities in $\mathcal{L}_{i}$, we use a set of meta summary entities $\mathcal{X}$ to guide the LLM to generate the summary entities. Here, $\mathcal{X}$ is a small set of general concepts such as "organization", "person", "location", "event", "technology", etc., that are generated by LLM. For example, the meta summary "technology" could guide the LLM to generate summary entities such as "big data" and "AI". Note that conceptually $\mathcal{X}$ is added as the top layer in Figure~\ref{fig:hirag_architecture}, but $\mathcal{X}$ is actually not part of the hierarchical KG.

After generating the summary entities and relations in the i-th layer, we update the KG as follows:
\begin{equation}
\mathcal{E}_{i} = \mathcal{E}_{i-1} \cup \mathcal{E}_{\{i-1,i\}},
\label{eq:10}
\end{equation}
\begin{equation}
\mathcal{V}_{i} = \mathcal{V}_{i-1} \cup \mathcal{L}_{i},
\label{eq:11}
\end{equation}
\begin{equation}
\mathcal{G}_{i} = \{(h,r,t)| h, t \in \mathcal{V}_{i}, r \in \mathcal{E}_{i}\}.
\label{eq:12}
\end{equation}

We repeat the above process for each layer from the 1st layer to the $k$-th layer. We will discuss how to choose the parameter $k$ in Section~\ref{sec:experiments}. Also note that there is no relation between the summary entities in each layer except the 0-th layer (i.e., the basic KG).



We also employ the Leiden algorithm~\cite{traag2019louvain} to compute a set of communities $\mathcal{P}$ from the hierarchical KG. Each community may contain entities from multiple layers and an entity may appear in multiple communities. 
For each community  $p \in \mathcal{P}$, we generate an interpretable semantic report using LLMs. Unlike existing methods such as GraphRAG~\cite{edge2024local} and LightRAG~\cite{guo2024lightrag}, which identify communities based solely on direct structural proximity in a basic KG, our hierarchical KG introduces multi-resolution semantic aggregation. Higher-layer entities in our KG act as semantic hubs that abstract clusters of semantically related entities regardless of their distance from each other in a lower layer. For example, while a flat KG might separate "cardiologist" and "neurologist" nodes due to limited direct connections, their hierarchical abstraction as "medical specialists" in upper layers enables joint community membership. The hierarchical structure thus provides dual connectivity enhancement: structural cohesion through localized lower-layer connections and semantic bridging via higher-layer abstractions. This dual mechanism ensures our communities reflect both explicit relational patterns and implicit conceptual relationships, yielding more comprehensive knowledge groupings than structure-only approaches. 


\subsection{Retrieval with Hierarchical Knowledge} \label{sec:hiretrieval}

We now discuss how we retrieve hierarchical knowledge to address the knowledge gap issue. Based on the hierarchical KG $\mathcal{G}_{k}$ constructed in Section~\ref{sec:hiindex}, we retrieve three-level knowledge at both local and global levels, as well as the bridging knowledge that connects them.

To retrieve local-level knowledge, we extract the top-$n$ most relevant entities $\hat{\mathcal{V}}$ as shown in Equation~\ref{eq:13}, where $\text{Sim}(q, v)$ is a function that measures the semantic similarity between a user query $q$ and an entity $v$ in the hierarchical KG $\mathcal{G}_{k}$. We set $n$ to 20 as default. 
\begin{equation}
\hat{\mathcal{V}} = \text{TopN}(\{v \in \mathcal{V}_{k} | \text{Sim}(q, v)\}, n).
\label{eq:13}
\end{equation}
To access global-level knowledge related to a query, we find the communities $\hat{\mathcal{P}} \subset \mathcal{P}$ that are connected to the retrieved entities as described in Equation~\ref{eq:14}, where $\mathcal{P}$ is computed during indexing in Section~\ref{sec:hiindex}. Then the community reports of these communities are retrieved, which represent coarse-grained knowledge relevant to the user's query.
\begin{equation}
\hat{\mathcal{P}} = \bigcup_{p \in \mathcal{P}}\{p | p \cap \hat{\mathcal{V}} \neq \emptyset\}.
\label{eq:14}
\end{equation}
To bridge the knowledge gap between the retrieved local-level and global-level knowledge, we also find a set of reasoning paths $\mathcal{R}$ connecting the retrieved communities. 
Specifically, from each community, we select the top-$m$ query-related key entities and collect them into $\hat{\mathcal{V}}_{\hat{\mathcal{P}}}$, as shown in Equation~\ref{eq:15}. 
The set of reasoning paths $\mathcal{R}$ is defined as the set of shortest paths between each pair of key entities according to their order in $\hat{\mathcal{V}}_{\hat{\mathcal{P}}}$, as shown in Equation~\ref{eq:16}. Based on $\mathcal{R}$, we construct a subgraph $\hat{\mathcal{R}}$ as described in Equation~\ref{eq:17}. Here, $\hat{\mathcal{R}}$ collects a set of triples from the KG that connect the knowledge in the local entities and the knowledge in the global communities.
\begin{equation}
\hat{\mathcal{V}}_{\hat{\mathcal{P}}} = \bigcup_{p \in \hat{\mathcal{P}}}\text{TopN}(\{v\in p | \text{Sim}(q, v)\}, m),
\label{eq:15}
\end{equation}
\begin{equation}
\mathcal{R} = \bigcup_{i\in [1, |\hat{\mathcal{V}}_{\hat{\mathcal{P}}}|-1]}\text{ShortestPath}_{\mathcal{G}_{k}}(\hat{\mathcal{V}}_{\hat{\mathcal{P}}}[i],\hat{\mathcal{V}}_{\hat{\mathcal{P}}}[i+1]),
\label{eq:16}
\end{equation}
\begin{equation}
\hat{\mathcal{R}} = 
\{(h,r,t) \in \mathcal{G}_{k}|h, t \in \mathcal{R}\}.
\label{eq:17}
\end{equation}
After retrieving the three-level hierarchical knowledge, i.e., local-level descriptions of the individual entities in $\hat{\mathcal{V}}$, global-level community reports of the communities in $\hat{\mathcal{P}}$, and bridge-level descriptions of the triples in $\hat{\mathcal{R}}$, we feed them as the context to the LLM to generate a comprehensive answer to the query. We also provide the detailed procedures of HiRAG with pseudocodes in Appendix~\ref{sec:appendix_hirag}.



\subsection{Why is HiRAG effective?} \label{sec:effective}

HiRAG’s efficacy stems from its hierarchical architecture, HiIndex (i.e., hierarchical KG) and HiRetrieval (i.e., three-level knowledge retrieval), which directly mitigates the limitations outlined in Challenges (1) and (2) as described in Section 1.

\textbf{Addressing Challenge (1):} The hierarchical knowledge graph $\mathcal{G}_{k}$ introduces summary entities in its higher layers, creating shortcuts between entities that are distantly located in lower layers. This design bridges semantically related concepts efficiently, bypassing the need for exhaustive traversal of fine-grained relationships in the KG.

\textbf{Resolving Challenge (2):} HiRetrieval constructs reasoning paths by linking the top-$n$ entities most semantically relevant to a query with their associated communities. These paths represent the shortest connections between localized entity descriptions and global community-level insights, ensuring that both granular details and broader contextual knowledge inform the reasoning process.

\textbf{Synthesis:} By integrating (i) semantically similar entities via hierarchical shortcuts, (ii) global community contexts, and (iii) optimized pathways connecting local and global knowledge, HiRAG generates comprehensive, context-aware answers to user queries.

\section{Experimental Evaluation}\label{sec:experiments}

We report the performance evaluation results of HiRAG in this section. 

\textbf{Baseline Methods.} We compared HiRAG with state-of-the-art and popular baseline RAG methods. \textbf{NaiveRAG}~\cite{gao2022precise, gao2023retrieval} splits original documents into chunks and retrieves relevant text chunks through vector search. \textbf{GraphRAG}~\cite{edge2024local} utilizes communities and we use the local search mode in our experiments as it retrieves community reports as global knowledge, while their global search mode is known to be too costly and does not use local entity descriptions. \textbf{LightRAG}~\cite{guo2024lightrag} uses both global and local knowledge to answer a query. \textbf{FastGraphRAG}~\cite{FastGraphRAG} integrates KG and personalized PageRank as proposed in HippoRAG~\cite{gutierrez2024hipporag}. \textbf{KAG}~\cite{liang2024kagboostingllmsprofessional} integrates structured reasoning of KG with LLMs and employs mutual indexing and logical-form-guided reasoning to enhance professional domain knowledge services.

\textbf{Datasets and Queries.} We used four datasets from the UltraDomain benchmark~\cite{qian2024memorag}, which is designed to evaluate RAG systems across diverse applications, focusing on long-context tasks and high-level queries in specialized domains. We used Mix, CS, Legal, and Agriculture datasets like in  LightRAG~\cite{guo2024lightrag}. We also used the benchmark queries provided in UltraDomain for each of the four datasets. The statistics of these datasets are given in Appendix~\ref{sec:appendix_dataset}.


\textbf{LLM.} We employed DeepSeek-V3~\cite{deepseekai2024deepseekv3technicalreport} as the LLM for information extraction, entity summarization, and answer generation in HiRAG and other baseline methods. We utilized GLM-4-Plus~\cite{glm2024chatglmfamilylargelanguage} as the embedding model for vector search and semantic clustering because DeepSeek-V3 does not provide an accessible embedding model.

\begin{table*}[t]
\centering
\caption{Win rates (\%) of HiRAG, its two variants (for ablation study), and baseline methods.}
\label{tab:performance}
\vspace{-0.1in}
\resizebox{\textwidth}{!}{
\begin{tabular}{@{}lcccccccc@{}}
\toprule
\textbf{}    & \multicolumn{2}{c}{\textbf{Mix}} & \multicolumn{2}{c}{\textbf{CS}} & \multicolumn{2}{c}{\textbf{Legal}} & \multicolumn{2}{c}{\textbf{Agriculture}} \\ 
\midrule
                & NaiveRAG & \textbf{HiRAG}& NaiveRAG & \textbf{HiRAG} & NaiveRAG & \textbf{HiRAG}& NaiveRAG & \textbf{HiRAG} \\
\cmidrule(lr){2-3}  \cmidrule(lr){4-5} \cmidrule(lr){6-7} \cmidrule(lr){8-9} 
Comprehensiveness& 16.6\%& \underline{83.4\%}& 30.0\%& \underline{70.0\%}& 32.5\%& \underline{67.5\%}& 34.0\%& \underline{66.0\%}\\
Empowerment& 11.6\%& \underline{88.4\%}& 29.0\%& \underline{71.0\%}& 25.0\%& \underline{75.0\%}& 31.0\%&\underline{69.0\%}\\
Diversity& 12.7\%& \underline{87.3\%}& 14.5\%& \underline{85.5\%}& 22.0\%& \underline{78.0\%}& 21.0\%&\underline{79.0\%}\\
Overall& 12.4\%& \underline{87.6\%}& 26.5\%& \underline{73.5\%}& 25.5\%& \underline{74.5\%}& 28.5\%&\underline{71.5\%}\\
\midrule
                & GraphRAG & \textbf{HiRAG}& GraphRAG & \textbf{HiRAG} & GraphRAG & \textbf{HiRAG}& GraphRAG & \textbf{HiRAG}\\
\cmidrule(lr){2-3}  \cmidrule(lr){4-5} \cmidrule(lr){6-7} \cmidrule(lr){8-9} 
Comprehensiveness& 42.1\%& \underline{57.9\%}& 40.5\%& \underline{59.5\%}& 48.5\%& \underline{51.5\%}& 49.0\%&\underline{51.0\%}\\
Empowerment& 35.1\%& \underline{64.9\%}& 38.5\%& \underline{61.5\%}& 43.5\%& \underline{56.5\%}& 48.5\%&\underline{51.5\%}\\
Diversity& 40.5\%& \underline{59.5\%}& 30.5\%& \underline{69.5\%}& 47.0\%& \underline{53.0\%}& 45.5\%&\underline{54.5\%}\\
Overall& 35.9\%& \underline{64.1\%}& 36.0\%& \underline{64.0\%}& 45.5\%& \underline{54.5\%}& 46.0\%&\underline{54.0\%}\\
\midrule
                & LightRAG & \textbf{HiRAG}& LightRAG & \textbf{HiRAG} & LightRAG & \textbf{HiRAG}& LightRAG & \textbf{HiRAG}\\
\cmidrule(lr){2-3}  \cmidrule(lr){4-5} \cmidrule(lr){6-7} \cmidrule(lr){8-9} 
Comprehensiveness& 36.8\%& \underline{63.2\%}& 44.5\%& \underline{55.5\%}& 49.0\%& \underline{51.0\%}& 38.5\%&\underline{61.5\%}\\
Empowerment& 34.9\%& \underline{65.1\%}& 41.5\%& \underline{58.5\%}& 43.5\%& \underline{56.5\%}& 36.5\%&\underline{63.5\%}\\
Diversity& 34.1\%& \underline{65.9\%}& 33.0\%& \underline{67.0\%}& \underline{63.0\%}& 37.0\%& 37.5\%&\underline{62.5\%}\\
Overall& 34.1\%& \underline{65.9\%}& 41.0\%& \underline{59.0\%}& 48.0\%& \underline{52.0\%}& 38.5\%&\underline{61.5\%}\\
\midrule
                & FastGraphRAG & \textbf{HiRAG}& FastGraphRAG & \textbf{HiRAG} & FastGraphRAG & \textbf{HiRAG}& FastGraphRAG & \textbf{HiRAG}\\
\cmidrule(lr){2-3}  \cmidrule(lr){4-5} \cmidrule(lr){6-7} \cmidrule(lr){8-9} 
Comprehensiveness& 0.8\%& \underline{99.2\%}& 0.0\%& \underline{100.0\%}& 1.0\%& \underline{99.0\%}& 0.0\%&\underline{100.0\%}\\
Empowerment& 0.8\%& \underline{99.2\%}& 0.0\%& \underline{100.0\%}& 0.0\%& \underline{100.0\%}& 0.0\%&\underline{100.0\%}\\
Diversity& 0.8\%& \underline{99.2\%}& 0.5\%& \underline{99.5\%}& 1.5\%& \underline{98.5\%}& 0.0\%&\underline{100.0\%}\\
Overall& 0.8\%& \underline{99.2\%}& 0.0\%& \underline{100.0\%}& 0.0\%& \underline{100.0\%}& 0.0\%&\underline{100.0\%}\\
\midrule
                & KAG & \textbf{HiRAG}& KAG & \textbf{HiRAG} & KAG & \textbf{HiRAG}& KAG & \textbf{HiRAG}\\
\cmidrule(lr){2-3}  \cmidrule(lr){4-5} \cmidrule(lr){6-7} \cmidrule(lr){8-9} 
Comprehensiveness& 2.3\%& \underline{97.7\%}& 1.0\%& \underline{99.0\%}& 16.5\%& \underline{83.5\%}& 5.0\%&\underline{99.5\%}\\
Empowerment& 3.5\%& \underline{96.5\%}& 4.5\%& \underline{95.5\%}& 9.0\%& \underline{91.0\%}& 5.0\%&\underline{99.5\%}\\
Diversity& 3.8\%& \underline{96.2\%}& 5.0\%& \underline{95.0\%}& 11.0\%& \underline{89.0\%}& 3.5\%&\underline{96.5\%}\\
Overall& 2.3\%& \underline{97.7\%}& 1.5\%& \underline{98.5\%}& 8.5\%& \underline{91.5\%}& 0.0\%&\underline{100.0\%}\\
\hhline{=========}
                & w/o HiIndex& \textbf{HiRAG}& w/o HiIndex& \textbf{HiRAG} & w/o HiIndex& \textbf{HiRAG} & w/o HiIndex& \textbf{HiRAG}\\
\cmidrule(lr){2-3}  \cmidrule(lr){4-5} \cmidrule(lr){6-7} \cmidrule(lr){8-9} 
Comprehensiveness& 46.7\%& \underline{53.3\%}& 44.2\%& \underline{55.8\%} & 49.0\%& \underline{51.0\%}& \underline{50.5\%}& 49.5\%\\
Empowerment& 43.2\%& \underline{56.8\%}& 38.8\%& \underline{61.2\%} & 47.5\%& \underline{52.5\%}& \underline{50.5\%}& 49.5\%\\
Diversity& 40.5\%& \underline{59.5\%}& 40.0\%& \underline{60.0\%} & 48.0\%& \underline{52.0\%}& 48.5\%& \underline{51.5\%}\\
Overall& 42.4\%& \underline{57.6\%}& 40.0\%& \underline{60.0\%} & 46.5\%& \underline{53.5\%}& 48.0\%& \underline{52.0\%}\\
\midrule
                & w/o Bridge& \textbf{HiRAG}& w/o Bridge& \textbf{HiRAG} & w/o Bridge& \textbf{HiRAG} & w/o Bridge& \textbf{HiRAG}\\
\cmidrule(lr){2-3}  \cmidrule(lr){4-5} \cmidrule(lr){6-7} \cmidrule(lr){8-9} 
Comprehensiveness& 49.2\%& \underline{50.8\%}& 46.5\%& \underline{53.5\%}& 49.5\%& \underline{50.5\%}& 47.0\%&\underline{53.0\%}\\
Empowerment& 44.2\%& \underline{55.8\%}& 43.0\%& \underline{57.0\%}& 38.5\%& \underline{61.5\%}& 41.0\%& \underline{59.0\%}\\
Diversity& 44.6\%& \underline{55.4\%}& 44.0\%& \underline{56.0\%}& 43.5\%& \underline{56.5\%}& 46.0\%& \underline{54.0\%}\\
Overall& 47.3\%& \underline{52.7\%}& 42.5\%& \underline{57.5\%}& 44.0\%& \underline{56.0\%}& 42.0\%& \underline{58.0\%}\\
\bottomrule
\end{tabular}
}
\end{table*}

\subsection{Overall Performance Comparison}\label{sec:main_result}

\textbf{Evaluation Details.} Our experiments followed the evaluation methods of recent work~\cite{edge2024local, guo2024lightrag} by employing a powerful LLM to conduct multi-dimensional comparison. We used the \textbf{win rate} to compare different methods, which indicates the percentage of instances that a method generates higher-quality answers compared to another method as judged by the LLM. We utilized GPT-4o~\cite{achiam2023gpt} as the evaluation model to judge which method generates a superior answer for each query for the following four dimensions: (1)~\textbf{Comprehensiveness}: how thoroughly does the answer address the question, covering all relevant aspects and details? (2)~\textbf{Empowerment}: how effectively does the answer provide actionable insights or solutions that empower the user to take meaningful steps? (3)~\textbf{Diversity}: how well does the answer incorporate a variety of perspectives, approaches, or solutions to the problem? (4)~\textbf{Overall}: how does the answer perform overall, considering comprehensiveness, empowerment, diversity, and any other relevant factors? For a fair comparison, we also alternated the order of the answers generated by each pair of methods in the prompts and calculated the overall win rates of each method.



\textbf{Evaluation Results.}  We present the win rates of HiRAG and five baseline methods in Table~\ref{tab:performance}. HiRAG consistently outperforms existing approaches across all four datasets and four evaluation dimensions in the majority of cases. Key insights derived from the results are summarized below.

\textit{Graph structure enhances RAG systems:} NaiveRAG exhibits inferior performance compared to methods integrating graph structures, primarily due to its inability to model relationships between entities in retrieved components. Furthermore, its context processing is constrained by the token limitations of LLMs, highlighting the importance of structured knowledge representation for robust retrieval and reasoning.

\textit{Global knowledge improves answer quality:} Approaches incorporating global knowledge (GraphRAG, LightRAG, KAG, HiRAG) significantly surpass FastGraphRAG, which relies on local knowledge via personalized PageRank. Answers generated without global context lack depth and diversity, underscoring the necessity of holistic knowledge integration for comprehensive responses.

\textit{HiRAG’s superior performance:} Among graph-enhanced RAG systems, HiRAG achieves the highest performance across all datasets (spanning diverse domains) and evaluation dimensions. This superiority stems primarily from two innovations: (1)~HiIndex which enhances connections between remote but semantically similar entities in the hierarchical KG, and (2) HiRetrieval which effectively bridges global knowledge with localized context to optimize relevance and coherence.

\subsection{Hierarchical KG vs. Flat KG}\label{section:hvf}

To evaluate the significance of the hierarchical KG, we replace the hierarchical KG with a flat KG (or a basic KG), denoted by \textbf{w/o HiIndex} as reported in Table~\ref{tab:performance}. Compared with HiRAG, the win rates of \textbf{w/o HiIndex} drop in almost all cases and quite significantly in at least half of the cases. This ablation study thus shows that the hierarchical indexing plays an important role in the quality of answer generation, since the connectivity among semantically similar entities is enhanced with the hierarchical KG, with which related entities can be grouped together both from structural and semantical perspectives. 

From Table~\ref{tab:performance}, we also observe that the win rates of \textbf{w/o HiIndex} are better or comparable to those of GraphRAG and LightRAG when compared with HiRAG. This suggests that our three-level knowledge retrieval method, i.e., HiRetrieval, is effective even applied on a flat KG, because GraphRAG and LightRAG also index on a flat KG but they only use the local entity descriptions and global community reports, while \textbf{w/o HiIndex} uses an additional bridge-level knowledge.



\begin{figure*}[htbp]
\vspace{-0.1in}
\centerline{\includegraphics[width=0.8\linewidth]{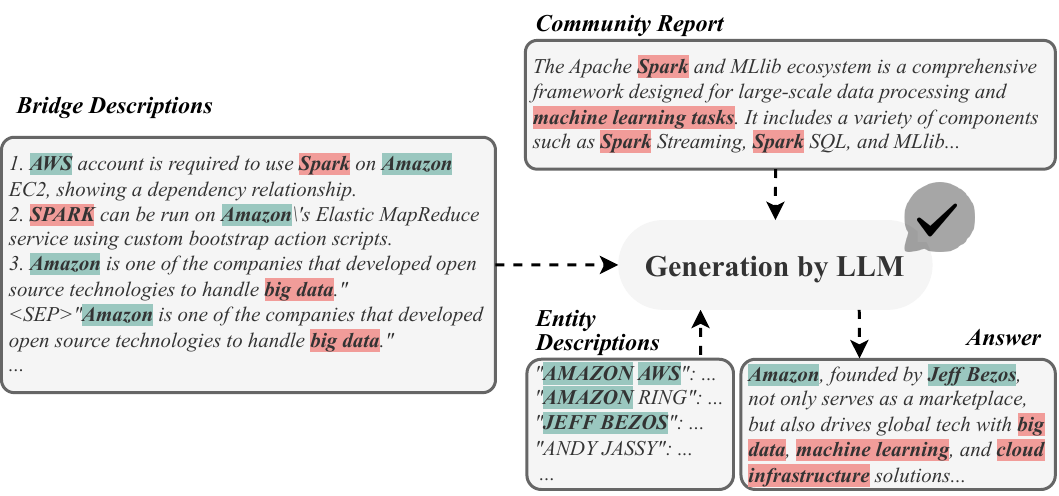}}
\caption{Answer to the query in Figure~\ref{fc} with additional bridge-level knowledge.}
\label{bridge}
\end{figure*}

\begin{table}[h]
    \setlength\tabcolsep{1pt}
    \centering
    \resizebox{0.47\textwidth}{!}{
          \begin{tabular}{ccccc}
            \toprule
            & \textbf{Mix}&\textbf{CS}  &\textbf{Legal} & \textbf{Agriculture}\\
            
            & Recall (\%)& Recall (\%)& Recall (\%)&Recall (\%)\\
            \midrule
            Global & 38.61& 48.96& 53.44&50.75 \\
            Local & 83.07&  81.88& 78.13&64.47 \\
            \bottomrule
        \end{tabular}
    }
\caption{The average knowledge coverage of bridge-level descriptions over global- and local-level knowledge across four datasets.}
\label{tab:proportion_bridge}
\vspace{-0.1in}
\end{table}
\subsection{HiRetrieval vs. Gapped Knowledge}\label{section:hvd}
To show the effectiveness of HiRetrieval, we also created another variant of HiRAG without using the bridge-level knowledge, denoted by \textbf{w/o Bridge} in Table~\ref{tab:performance}. The result shows that without the bridge-layer knowledge, the win rates drop significantly across all datasets and evaluation dimensions, because there is knowledge gap between the local-level and global-level knowledge as discussed in Section~\ref{sec:intro}. We also report the knowledge coverage of bridge-level descriptions over global- and local-level knowledge in Table~\ref{tab:proportion_bridge} and Appendix~\ref{appendix:proportion_bridge}, which further proofs both local- and global-level knowledge are well connected in the bridge-level descriptions.



\textbf{Case Study.} Figure~\ref{bridge} shows the three-level knowledge used as the context to an LLM to answer the query in Figure~\ref{fc}.  The bridge-level knowledge contains entity descriptions from different communities, as shown by the different colors in Figure~\ref{bridge}, which helps the LLM correctly answer the question about Amazon’s role as an e-commerce and cloud provider.



\subsection{Determining the Number of Layers}\label{section:layers}
\begin{figure}[htbp]
\centerline{\includegraphics[width=0.75\linewidth]{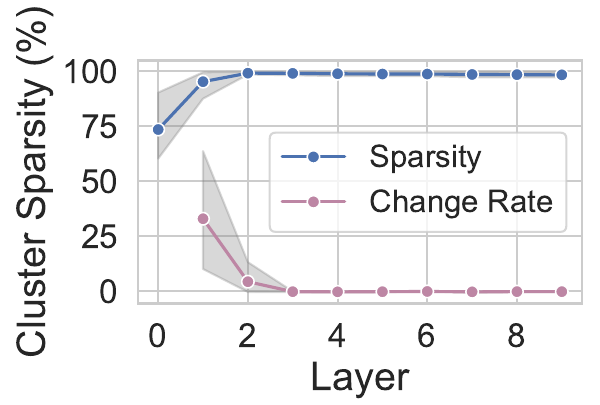}}
\caption{Cluster sparsity $CS_{i}$ and change rate from $CS_{i}$ to $CS_{i+1}$, where the shadow areas represent the value ranges of the four datasets and the blue/pink lines are the respective average values.}
\label{cs}
\vspace{-0.1in}
\end{figure}

\begin{table*}[t]\centering
    \vspace{-0.1in}
    \resizebox{0.85\textwidth}{!}{
    \Large
    \begin{tabular}{*{1}{l}|*{1}{c}|*{2}{c}|*{2}{c}|*{2}{c}}
        \toprule
         \multicolumn{1}{c|}{} & \multicolumn{1}{c|}{} & \multicolumn{2}{c|}{Token Cost} & \multicolumn{2}{c|}{API Calls} & \multicolumn{2}{c}{Time Cost (s)} \\
          Dataset & Method & Indexing & Retrieval  & Indexing & Retrieval & Indexing & Retrieval \\
        \midrule
        \multirow{4}{*}{\textbf{Mix}} & GraphRAG  & 8,507,697 & 0.00 & 2,666 & 1.00 & 6,696 & 0.70 \\ %
        & LightRAG  & 3,849,030 & 357.76 & 1,160 & 2.00 & 3,342 & 3.06  \\
        & KAG  & 6,440,668 & 110,532.00 & 831 & 9.17 & 8,530 & 58.47 \\
        & HiRAG & 21,898,765 & 0.00 & 6,790 & 1.00 & 17,208 & 0.85 \\ %
        \midrule
        \multirow{4}{*}{\textbf{CS}} & GraphRAG  & 27,506,689 & 0.00 & 8,649 & 1.00 & 19,255 & 0.98  \\ %
        & LightRAG  & 12,638,997 & 353.37 & 3,799 & 2.00 & 14,307 & 4.97  \\
        & KAG  & 7,358,717 & 89,746.00 & 2,190 & 6.29 & 14,837 & 46.37 \\
        & HiRAG & 56,042,906 & 0.00 & 16,535 & 1.00 & 44,994 & 1.17 \\ %
        \midrule
        \multirow{4}{*}{\textbf{Legal}} & GraphRAG  & 51,168,359 & 0.00 & 13,560 & 1.00 & 30,065 & 1.12 \\ %
        & LightRAG & 30,299,958 & 353.77 & 9,442 & 2.00 & 21,505 & 5.44  \\
        & KAG  & 18,431,706 & 97,683.00 & 4,980 & 7.82 & 29,191 & 51.26 \\
        & HiRAG & 106,427,778 & 0.00 & 27,224 & 1.00 & 115,232 & 2.04 \\ %
        \midrule
        \multirow{4}{*}{\textbf{Agriculture}} & GraphRAG  & 27,974,472 & 0.00 & 8,669 & 1.00 & 20,362 & 1.17 \\ %
        & LightRAG  & 12,031,096 & 354.62 & 3,694 & 2.00 & 13,550 & 5.64  \\
        & KAG  & 7,513,424 & 93,217.00 & 2,358 & 6.83 & 22,557 & 49.57 \\
        & HiRAG & 96,080,883 & 0.00 & 22,736 & 1.00 & 50,920 & 1.76 \\ %
        \bottomrule
    \end{tabular}
    }
\caption{Comparisons in terms of tokens, API calls and time cost across four datasets.} %
\label{tab:cost_study}
\vspace{-0.1in}
\end{table*}

One important thing in HiIndex is to determine the number of layers, $k$, for the hierarchical KG, which should be determined dynamically according to the quality of clusters in each layer. We stop building another layer when the majority of the clusters consist of only a small number of entities, meaning that the entities can no longer be effectively grouped together. To measure that, we introduce the notion of \textbf{cluster sparsity} $CS_i$, as inspired by graph sparsity, to measure the quality of clusters in the $i$-th layer as described in Equation~\ref{eq:19}. 
\begin{equation}
CS_i = 1 - \frac{\sum_{\mathcal{S}\in \mathcal{C}_{i}}|\mathcal{S}|(|\mathcal{S}|-1)}{|\mathcal{L}_{i}|(|\mathcal{L}_{i}|-1)}.
\label{eq:19}
\end{equation}
The more the clusters in $\mathcal{C}_{i}$ have a small number of entities, the larger is $CS_i$, where the worst case is when each cluster contains only one entity (i.e., $CS_i=1$).  Figure~\ref{cs} shows that as we have more layers, the cluster sparsity increases and then stabilizes. We also plot the change rate from $CS_{i}$ to $CS_{i+1}$, which shows that there is little or no more change after constructing a certain number of layers. We set a threshold $\epsilon = 5\%$  and stop constructing another layer when the change rate of cluster sparsity is lower than $\epsilon$ because the cluster quality has little or no improvement after that.

\subsection{Efficiency and Costs Analysis}\label{sec:cost}
To evaluate the efficiency and costs of HiRAG, we also report the token costs, the number of API calls, and the time costs of indexing and retrieval of HiRAG and the baselines in Table~\ref{tab:cost_study}. For indexing, we record the total costs of the entire indexing process. For retrieval, we calculate the average costs per query during the retrieval process, which could reflect the performance while the methods are deployed online. 

Although HiRAG needs more time and resources to conduct indexing for better performance, we remark that indexing is offline and the total cost is only about 7.55 USD for the Mix dataset using DeepSeek-V3.  In terms of retrieval, unlike KAG and LightRAG, HiRAG does not cost any tokens for retrieval. Therefore, HiRAG is more efficient for online retrieval.

\section{Conclusions}
We presented a new approach to enhance RAG systems by effectively utilizing graph structures with hierarchical knowledge. By developing (1)~HiIndex which enhances structural and semantic connectivity across hierarchical layers, and (2)~HiRetrieval which effectively bridges global conceptual abstractions with localized entity descriptions, HiRAG achieves superior performance than existing methods.

\section{Limitations}
HiRAG has the following limitations. Firstly, constructing a high-quality hierarchical KG may incur substantial token consumption and time overhead, as LLMs need to perform entity summarization in each layer. However, the monetary cost of using LLMs may not be the major concern as the cost is decreasing rapidly recently, and therefore we may consider parallelizing the indexing process to reduce the indexing time. Secondly, the retrieval module requires more sophisticated query-aware ranking mechanisms. Currently, our HiRetrieval module relies solely on LLM-generated weights for relation ranking, which may affect query relevance. In the future, we will research for more effective ranking mechanisms to further improve the retrieval quality. Besides, we can also incorporate causality into HiRAG to enhance reasoning capabilities of LLMs~\cite{liu2024discovery, chen2022learning}.




\bibliography{acl_latex}

\appendix
\section*{Appendix}\label{sec:appendix}
\section{Experimental Datasets}\label{sec:appendix_dataset}
\begin{table}[thb]
	\setlength\tabcolsep{6pt}
	\renewcommand{\arraystretch}{1.2}
	\centering
	\resizebox{0.48\textwidth}{!}{
		\begin{tabular}{lllll}
			\hline
			\textbf{Dataset}& \textbf{Mix}& \textbf{CS}&\textbf{Legal}&\textbf{Agriculture}\\
			\hline
			 \# of Documents& 61& 10&94&12\\
			\# of Tokens& 625948&2210894&5279400&2028496\\
                \hline
		\end{tabular}
	}
\caption{Statistics of datasets. }
\label{tab:statistics}
\end{table}
Table~\ref{tab:statistics} presents the statistical characteristics of the experimental datasets, where all documents were consistently tokenized using Byte Pair Encoding (BPE) tokenizer "cl100k\_base".

\section{Evaluations with Objective Metrics}
\begin{table}[htb]
	\setlength\tabcolsep{6pt}
	\renewcommand{\arraystretch}{1.2}
	\centering
\resizebox{0.48\textwidth}{!}{
      \begin{tabular}{ccccc}
        \toprule
        & \multicolumn{2}{c}{2WikiMultiHopQA}& \multicolumn{2}{c}{HotpotQA}\\
        
        Method& EM (\%)& F1 (\%)& EM (\%)&F1 (\%)\\
        \midrule
        NaiveRAG & 15.60& 25.64& 21.60&40.19\\
        GraphRAG  & 22.50& 27.49& 31.70&42.74\\
        LightRAG & 16.50&40.95& 25.00&43.20\\
        FastGraphRAG & 20.80&44.81& 35.00&49.56\\
        \midrule
        HiRAG &  \textbf{46.20}&  \textbf{60.06}& \textbf{37.00}&\textbf{52.29}\\
        \bottomrule
    \end{tabular}
}
\caption{QA performances of HiRAG and other baseline methods with EM and F1 scores. }
\label{tab:QAbench}
\end{table}
To objectively evaluate the QA performance of HiRAG and the baseline methods, we leverage two established metrics: exact match (\textbf{EM}) and \textbf{F1} scores, applied to the generated answers. We perform systematic evaluations using GPT-4o-mini on two multi-hop QA datasets: \textbf{HotpotQA}~\cite{yang2018hotpotqa} and \textbf{2WikiMultiHopQA}~\cite{ho2020constructing}. For a consistent comparison with previous work, we follow the settings of HippoRAG~\cite{gutierrez2024hipporag}, obtaining 1,000 queries from each validation set. We did not present the results of KAG because, despite our efforts to implement it, we were unable to make it fully work on this benchmark. 

Compared with the metric of win rates, the performances with EM and F1 scores can indicate HiRAG's ability to achieve correctness. Given that the RAG system has access to richer contexts, it tends to produce more comprehensive responses. Nevertheless, while comprehensiveness, empowerment, and diversity are important qualities for the generated answers, correctness is equally essential. As illustrated in Table~\ref{tab:QAbench}, HiRAG is also capable of generating more accurate answers compared to the baseline methods.

\section{Implementation Details of HiRAG}\label{sec:appendix_hirag}
\begin{algorithm}[!h]
	\renewcommand{\algorithmicrequire}{\textbf{Input:}}
	\renewcommand{\algorithmicensure}{\textbf{Output:}}
	\caption{HiIndex}
    \label{algorithm:a1}
	\begin{algorithmic}[1]  
		\REQUIRE  Basic knowledge graph $\mathcal{G}_{0}$ extracted by the LLM; Predefined threshold $\epsilon$;
		\ENSURE Hierarchical knowledge graph $\mathcal{G}_{k}$;
        \STATE $\mathcal{L}_{0} \leftarrow \mathcal{V}_{0}$;
        \STATE $\mathcal{Z}_{0} \leftarrow \{\text{Embedding}(v) | v \in \mathcal{L}_{0}\}$;
        \STATE $i \leftarrow 1$;
		\WHILE{True}
        \STATE \textit{/*Perform semantical clustering*/}
        \STATE $\mathcal{C}_{i-1} \leftarrow \text{GMM}(\mathcal{G}_{i-1}, \mathcal{Z}_{i-1})$;
        \STATE \textit{/*Calculate cluster sparsity*/}
		\STATE $CS_{i} \leftarrow 1 - \frac{\sum_{\mathcal{S} \in \mathcal{C}_{i-1}}|\mathcal{S}|(|\mathcal{S}|-1)}{|\mathcal{L}_{i-1}|(|\mathcal{L}_{i-1}|-1)}$;
        \IF{change rate of $CS_{i}$ $ \leq \epsilon$}
        \STATE $i \leftarrow i - 1$;
        \STATE break;
        \ENDIF
        \STATE \textit{/*Generate summary entities and relations*/}
        \STATE $\mathcal{L}_{i} \leftarrow \{\}$;
        \STATE $\mathcal{E}_{\{i-1,i\}} \leftarrow \{\}$;
        \FOR{$\mathcal{S}_x $ in $\mathcal{C}_{i-1}$}
        \STATE $\mathcal{L}, \mathcal{E} \leftarrow  \text{LLM}(\mathcal{S}_{x}, \mathcal{X})$;
        \STATE $\mathcal{L}_{i} \leftarrow \mathcal{L}_{i} \cup \mathcal{L}$;
        \STATE $\mathcal{E}_{\{i-1,i\}} \leftarrow \mathcal{E}_{\{i-1,i\}} \cup \mathcal{E}$;
        \ENDFOR
        \STATE $\mathcal{Z}_{i} = \{\text{Embedding}(v)|v \in \mathcal{L}_{i}\}$;
        \STATE \textit{/*Update KG*/}
        \STATE $\mathcal{E}_{i} \leftarrow \mathcal{E}_{i-1} \cup \mathcal{E}_{\{i-1,i\}}$;
        \STATE $\mathcal{V}_{i} \leftarrow \mathcal{V}_{i-1} \cup \mathcal{L}_{i}$;
        \STATE $\mathcal{G}_{i}  \leftarrow \{(h,r,t)| h, t \in \mathcal{V}_{i}, r \in \mathcal{E}_{i}\}$
        \STATE $i \leftarrow i + 1$;
        \ENDWHILE
        \STATE $k \leftarrow i$;
        \STATE $\mathcal{G}_{k} \leftarrow \{(h,r,t)| h, t \in \mathcal{V}_{k}, r \in \mathcal{E}_{k}\}$;
	\end{algorithmic}
\end{algorithm}

\begin{algorithm}[!h]
	\renewcommand{\algorithmicrequire}{\textbf{Input:}}
	\renewcommand{\algorithmicensure}{\textbf{Output:}}
	\caption{HiRetrieval}
    \label{algorithm:a2}
	\begin{algorithmic}[1]  
		\REQUIRE  The hierarchical knowledge graph $\mathcal{G}_{k}$; The detected community set $\mathcal{P}$ in $\mathcal{G}_{k}$; The number of retrieved entities $n$; The number of selected key entities $m$ in each retrieved community;
		\ENSURE The generated answer $a$;
        \STATE \textit{/*The local-layer knowledge context*/}
        \STATE $\hat{\mathcal{V}} \leftarrow \text{TopN}(\{v \in \mathcal{V}_{k}|\text{Sim}(v,q)\}, n)$;
        \STATE \textit{/*The global-layer knowledge context*/}
        \STATE $\hat{\mathcal{P}} \leftarrow \bigcup_{p \in \mathcal{P}}\{p | p \cap \hat{\mathcal{V}} \neq \phi\}$;
        \STATE $\hat{\mathcal{R}} \leftarrow \{\}$;
        \STATE $\hat{\mathcal{V}}_{\hat{\mathcal{P}}} \leftarrow \{\}$;
        \STATE \textit{/*Select key entities*/}
        \FOR{$p$ in $\hat{\mathcal{P}}$}
        \STATE $\hat{\mathcal{V}}_{\hat{\mathcal{P}}} \leftarrow \hat{\mathcal{V}}_{\hat{\mathcal{P}}} \cup \text{TopN}(\{v \in p | \text{Sim}(v,q)\}, m)$;
        \ENDFOR
        \STATE \textit{/*Find the reasoning path*/}
        \FOR{$i$ in $[1, |\hat{\mathcal{V}}_{\hat{\mathcal{P}}}|-1]$}
        \STATE $\mathcal{R} \leftarrow \mathcal{R} \cup \text{ShortestPath}_{\mathcal{G}_{k}}(\hat{\mathcal{V}}_{\hat{\mathcal{P}}}[i],\hat{\mathcal{V}}_{\hat{\mathcal{P}}}[i+1])$;
        \ENDFOR
        \STATE \textit{/*The bridge-layer knowledge context*/}
        \STATE $\hat{\mathcal{R}} \leftarrow \{(h,r,t) \in \mathcal{G}_{k} | h,t \in \mathcal{R}\}$;
        \STATE \textit{/*Generate the answer*/}
        \STATE $a \leftarrow \text{LLM}(q, \hat{\mathcal{V}}, \hat{R}, \hat{\mathcal{P}})$;
	\end{algorithmic}
\end{algorithm}
We give a more detailed and formulated expression of hierarchical indexing (HiIndex) and hierarchical retrieval (HiRetrieval). As described in Algorithm~\ref{algorithm:a1}, the hierarchical knowledge graph is constructed iteratively. The number of clustered layers depends on the rate of change in the cluster sparsity at each layer. As shown in Algorithm~\ref{algorithm:a2}, we retrieve knowledge of three layers (local layer, global layer, and bridge layer) as contexts for LLM to generate more comprehensive and accurate answers.

\begin{table*}[!t]
\vspace{-0.1in}
\resizebox{\textwidth}{!}{
\begin{tabular}{@{}lcccccc@{}}
\toprule
\textbf{}    & \multicolumn{2}{c}{\textbf{GPT-4o}} & \multicolumn{2}{c}{\textbf{Claude-3.5-sonnet}} & \multicolumn{2}{c}{\textbf{Qwen-turbo}} \\ 
\midrule
                & NaiveRAG & \textbf{HiRAG}& NaiveRAG & \textbf{HiRAG} & NaiveRAG & \textbf{HiRAG} \\
\cmidrule(lr){2-3}  \cmidrule(lr){4-5} \cmidrule(lr){6-7}
Comprehensiveness& 16.6\%& \underline{83.4\%}& 13.0\%& \underline{87.0\%}& 13.6\%& \underline{86.4\%}\\
Empowerment& 11.6\%& \underline{88.4\%}& 10.0\%& \underline{90.0\%}& 12.7\%& \underline{87.3\%}\\
Diversity& 12.7\%& \underline{87.3\%}& 28.0\%& \underline{72.0\%}& 18.2\%& \underline{81.8\%}\\
Overall& 12.4\%& \underline{87.6\%}& 11.0\%& \underline{89.0\%}& 12.7\%& \underline{87.3\%}\\
\midrule
                & GraphRAG & \textbf{HiRAG}& GraphRAG & \textbf{HiRAG} & GraphRAG & \textbf{HiRAG}\\
\cmidrule(lr){2-3}  \cmidrule(lr){4-5} \cmidrule(lr){6-7} 
Comprehensiveness& 42.1\%& \underline{57.9\%}& 39.1\%& \underline{60.9\%}& 32.4\%& \underline{67.6\%}\\
Empowerment& 35.1\%& \underline{64.9\%}& 31.8\%& \underline{68.2\%}& 33.3\%& \underline{66.7\%}\\
Diversity& 40.5\%& \underline{59.5\%}& 48.2\%& \underline{51.8\%}& 40.7\%& \underline{59.3\%}\\
Overall& 35.9\%& \underline{64.1\%}& 32.7\%& \underline{67.3\%}& 32.4\%& \underline{67.6\%}\\
\midrule
                & LightRAG & \textbf{HiRAG}& LightRAG & \textbf{HiRAG} & LightRAG & \textbf{HiRAG}\\
\cmidrule(lr){2-3}  \cmidrule(lr){4-5} \cmidrule(lr){6-7} 
Comprehensiveness& 36.8\%& \underline{63.2\%}& 36.4\%& \underline{63.6\%}& 35.5\%& \underline{64.5\%}\\
Empowerment& 34.9\%& \underline{65.1\%}& 31.8\%& \underline{68.2\%}& 35.5\%& \underline{64.5\%}\\
Diversity& 34.1\%& \underline{65.9\%}& 40.1\%& \underline{59.1\%}& 39.1\%& \underline{60.9\%}\\
Overall& 34.1\%& \underline{65.9\%}& 33.6\%& \underline{66.4\%}& 35.5\%& \underline{64.5\%}\\
\midrule
                & FastGraphRAG & \textbf{HiRAG}& FastGraphRAG & \textbf{HiRAG} & FastGraphRAG & \textbf{HiRAG}\\
\cmidrule(lr){2-3}  \cmidrule(lr){4-5} \cmidrule(lr){6-7}
Comprehensiveness& 0.8\%& \underline{99.2\%}& 0.0\%& \underline{100.0\%}&0.8\%&\underline{99.2\%}\\
Empowerment& 0.8\%& \underline{99.2\%}& 0.0\%& \underline{100.0\%}& 0.8\%& \underline{99.2\%}\\
Diversity& 0.8\%& \underline{99.2\%}& 0.9\%& \underline{99.1\%}& 0.8\%& \underline{99.2\%}\\
Overall& 0.8\%& \underline{99.2\%}& 0.0\%& \underline{100.0\%}& 0.8\%& \underline{99.2\%}\\
\midrule
                & KAG & \textbf{HiRAG}& KAG & \textbf{HiRAG} & KAG & \textbf{HiRAG}\\
\cmidrule(lr){2-3}  \cmidrule(lr){4-5} \cmidrule(lr){6-7} 
Comprehensiveness& 2.3\%& \underline{97.7\%}& 1.8\%& \underline{98.2\%}& 3.6\%& \underline{96.4\%}\\
Empowerment& 3.5\%& \underline{96.5\%}& 2.7\%& \underline{97.3\%}& 5.5\%& \underline{94.5\%}\\
Diversity& 3.8\%& \underline{96.2\%}& 12.7\%& \underline{87.3\%}& 10.9\%& \underline{89.1\%}\\
Overall& 2.3\%& \underline{97.7\%}& 1.8\%& \underline{98.2\%}& 3.6\%& \underline{96.4\%}\\
\bottomrule
\end{tabular}
}
\caption{Win rates (\%) of HiRAG and baseline methods on four tasks with three different powerful LLMs as the evaluator.}
\label{tab:qfs_reuslt_cross}
\vspace{-0.1in}
\end{table*}

\section{The Clustering Coefficients of HiIndex}
\begin{figure}[htbp]
\centerline{\includegraphics[width=0.85\linewidth]{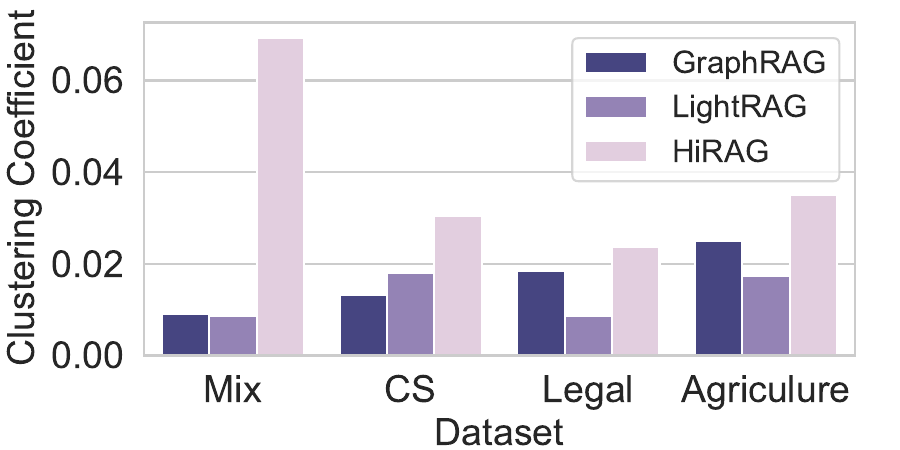}}
\caption{Comparisons between the clustering coefficients of GraphRAG, LightRAG and HiRAG across four datasets.}
\label{connect}
\end{figure}

We calculate and compare the clustering coefficients of GraphRAG, LightRAG and HiRAG in Figure~\ref{connect}. HiRAG shows a higher clustering coefficient than other baseline methods, which means that more entities in the hierarchical KG constructed by the HiIndex module tend to cluster together. And this is also the reason why the HiIndex module can improve the performance of RAG systems.

\section{A Simple Case of Hierarchical KG}
\begin{figure}[h]
\centerline{\includegraphics[width=0.9\linewidth]{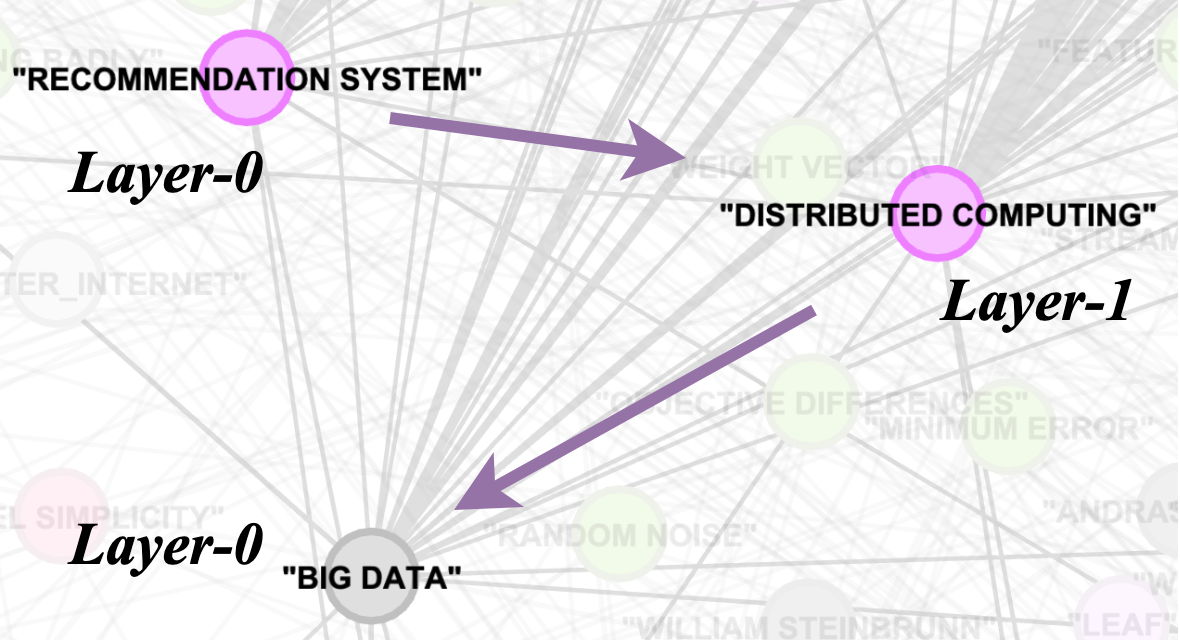}}
\caption{The shortest path with hierarchical KG between the entities in the case mentioned in the introduction.}
\label{fixcase}
\vspace{-0.15in}
\end{figure}
As shown in Figure~\ref{fixcase}, we fix the issues mentioned in Section~\ref{sec:intro} with a hierarchical KG. This case demonstrates that the GMMs clustered semantically similar entities "BIG DATA" and "RECOMMENDATION SYSTEM" together. The LLM summarizes "DISTRIBUTED COMPUTING" as their shared summary entities in the next layer. As a consequence, the connections between these related entities can be enhanced from a semantic perspective.

\section{Cross Validation for LLM as a Judge}\label{appendix:cv_gj}
Although our LLM-based evaluation approach for the query-focused summarization task is a common practice in the performance evaluation by existing graph RAG methods~\cite{es2024ragas, guo2024lightrag, edge2024local}, we also conducted cross-verification using \textbf{Qwen-turbo} and \textbf{Claude-3.5-sonnet} as the LLM judge to further make our experimental results more convincing. As shown in Table~\ref{tab:qfs_reuslt_cross}, the results consistently demonstrate HiRAG's superiority over all the other graph RAG methods compared, confirming that our conclusions remain stable even when neutralizing LLM evaluator-specific biases. To reduce the cost, we report the results on the Mix dataset while the results on the other datasets follow a similar pattern.

\section{Knowledge Coverage of Bridge-Level Descriptions}\label{appendix:proportion_bridge}
To further validate that both local- and global-level knowledge are well connected in the bridge-level descriptions, we counted the average token-level recall ratio of local- and global-level context in the bridge-level context across four datasets. As shown in Table~\ref{tab:proportion_bridge}, the results demonstrate that our bridge-level retrieval effectively captures a significant portion of both entity-level and community-level information, providing empirical support for the method's effectiveness. Here, the recall indicates that the percentage of global-level info or local-level info that is captured in the bridge-level context.

\section{Prompt Templates used in HiRAG}

\begin{figure*}[!h]
    \centering
    \scriptsize
    \begin{AIbox}{Entity Extraction}
    
    \begin{flushleft}
    \ttfamily
    -Goal-\\
    Given a text document that is potentially relevant to a list of entity types, identify all entities of those types.\\[4pt]
    
    -Steps-\\
    1. Identify all entities. For each identified entity, extract the following information:\\
    - entity\_name: Name of the entity, capitalized\\
    - entity\_type: One of the following types: [\{entity\_types\}], normal\_entity means that doesn't belong to any other types.\\
    - entity\_description: Comprehensive description of the entity's attributes and activities\\[2pt]
    Format each entity as (\string"entity\string"\{tuple\_delimiter\}\textless entity\_name\textgreater\{tuple\_delimiter\}\textless entity\_type\textgreater\{tuple\_delimiter\}\textless entity\_description\textgreater)\\
    Return output in English as a single list of all the entities identified in step 1. Use **\{record\_delimiter\}** as the list delimiter.\\[4pt]
    
    3. When finished, output \{completion\_delimiter\}\\[6pt]
    
    \#\#\#\#\#\#\#\#\#\#\#\#\#\#\#\#\#\#\#\#\#\#\\
    -Examples-\\
    \#\#\#\#\#\#\#\#\#\#\#\#\#\#\#\#\#\#\#\#\#\#\\
    Example 1:\\
    
    Entity\_types: [person, technology, mission, organization, location]\\
    Text:\\
    while Alex clenched his jaw, the buzz of frustration dull against the backdrop of Taylor's authoritarian certainty. It was this competitive undercurrent that kept him alert, the sense that his and Jordan's shared commitment to discovery was an unspoken rebellion against Cruz's narrowing vision of control and order.\\[2pt]
    
    Then Taylor did something unexpected. They paused beside Jordan and, for a moment, observed the device with something akin to reverence. ``If this tech can be understood\ldots'' Taylor said, their voice quieter, ``It could change the game for us. For all of us.''\\[2pt]
    
    The underlying dismissal earlier seemed to falter, replaced by a glimpse of reluctant respect for the gravity of what lay in their hands. Jordan looked up, and for a fleeting heartbeat, their eyes locked with Taylor's, a wordless clash of wills softening into an uneasy truce.\\[2pt]
    
    It was a small transformation, barely perceptible, but one that Alex noted with an inward nod. They had all been brought here by different paths\\
    \#\#\#\#\#\#\#\#\#\#\#\#\#\#\#\#\#\#\#\#\#\#\\
    Output:\\
    \upshape
    (\string"entity\string"\{tuple\_delimiter\}Alex\{tuple\_delimiter\}person\{tuple\_delimiter\}Alex is a character who experiences frustration and is observant of the dynamics among other characters.)\{record\_delimiter\}\\
    (\string"entity\string"\{tuple\_delimiter\}Taylor\{tuple\_delimiter\}person\{tuple\_delimiter\}Taylor is portrayed with authoritarian certainty and shows a moment of reverence towards a device, indicating a change in perspective.)\{record\_delimiter\}\\
    (\string"entity\string"\{tuple\_delimiter\}Jordan\{tuple\_delimiter\}person\{tuple\_delimiter\}Jordan shares a commitment to discovery and has a significant interaction with Taylor regarding a device.)\{record\_delimiter\}\\
    (\string"entity\string"\{tuple\_delimiter\}Cruz\{tuple\_delimiter\}person\{tuple\_delimiter\}Cruz is associated with a vision of control and order, influencing the dynamics among other characters.)\{record\_delimiter\}\\
    (\string"entity\string"\{tuple\_delimiter\}The Device\{tuple\_delimiter\}technology\{tuple\_delimiter\}The Device is central to the story, with potential game-changing implications, and is revered by Taylor.)\{record\_delimiter\}\\[6pt]
    
    \ttfamily
    \#\#\#\#\#\#\#\#\#\#\#\#\#\#\#\#\#\#\#\#\#\#\\
    Example 2:\\
    \ldots\\
    \#\#\#\#\#\#\#\#\#\#\#\#\#\#\#\#\#\#\#\#\#\#\\
    Example 3:\\
    \ldots\\
    \#\#\#\#\#\#\#\#\#\#\#\#\#\#\#\#\#\#\#\#\#\#\\
    
    -Real Data-\\
    \#\#\#\#\#\#\#\#\#\#\#\#\#\#\#\#\#\#\#\#\#\#\\
    Entity\_types: \{entity\_types\}\\
    Text: \{input\_text\}\\
    \#\#\#\#\#\#\#\#\#\#\#\#\#\#\#\#\#\#\#\#\#\#\\
    Output:
    \end{flushleft}
    
    \end{AIbox}
    \caption{The prompt template designed to extract entities from text chunks.}
    \label{prompt_ee}
\end{figure*}
    
    \subsection{Prompt Templates for Entity Extraction}~\label{appendix:prompt_ee}
    As shown in Figure~\ref{prompt_ee}, we used that prompt template to extract entities from text chunks. We also give three examples to guide the LLM to extract entities with higher accuracy.
    
    \begin{figure*}[!h]
    \centering
    \scriptsize
    \begin{AIbox}{Relation Extraction}
    
    \begin{flushleft}
    \ttfamily
    -Goal-\\
    Given a text document that is potentially relevant to a list of entities, identify all relationships among the given identified entities.\\[4pt]
    
    -Steps-\\
    1. From the entities given by user, identify all pairs of (source\_entity, target\_entity) that are *clearly related* to each other.\\
    For each pair of related entities, extract the following information:\\
    - source\_entity: name of the source entity, as identified in step 1\\
    - target\_entity: name of the target entity, as identified in step 1\\
    - relationship\_description: explanation as to why you think the source entity and the target entity are related to each other\\
    - relationship\_strength: a numeric score indicating strength of the relationship between the source entity and target entity\\[2pt]
    Format each relationship as (\string"relationship\string"\{tuple\_delimiter\}\textless source\_entity\textgreater\{tuple\_delimiter\}\textless target\_entity\textgreater\{tuple\_delimiter\}\textless \\relationship\_description\textgreater\{tuple\_delimiter\}\textless relationship\_strength\textgreater)\\[4pt]
    
    2. Return output in English as a single list of all the entities and relationships identified in steps 1 and 2. Use **\{record\_delimiter\}** as the list delimiter.\\[4pt]
    
    3. When finished, output \{completion\_delimiter\}\\[6pt]
    
    \#\#\#\#\#\#\#\#\#\#\#\#\#\#\#\#\#\#\#\#\#\#\\
    -Examples-\\
    \#\#\#\#\#\#\#\#\#\#\#\#\#\#\#\#\#\#\#\#\#\#\\
    Example 1:\\
    
    Entities: ["Alex", "Taylor", "Jordan", "Cruz", "The Device"]\\
    Text:\\
    while Alex clenched his jaw, the buzz of frustration dull against the backdrop of Taylor's authoritarian certainty. It was this competitive undercurrent that kept him alert, the sense that his and Jordan's shared commitment to discovery was an unspoken rebellion against Cruz's narrowing vision of control and order.\\[2pt]
    
    Then Taylor did something unexpected. They paused beside Jordan and, for a moment, observed the device with something akin to reverence. ``If this tech can be understood\ldots'' Taylor said, their voice quieter, ``It could change the game for us. For all of us.''\\[2pt]
    
    The underlying dismissal earlier seemed to falter, replaced by a glimpse of reluctant respect for the gravity of what lay in their hands. Jordan looked up, and for a fleeting heartbeat, their eyes locked with Taylor's, a wordless clash of wills softening into an uneasy truce.\\[2pt]
    
    It was a small transformation, barely perceptible, but one that Alex noted with an inward nod. They had all been brought here by different paths\\
    \#\#\#\#\#\#\#\#\#\#\#\#\#\#\#\#\#\#\#\#\#\#\\
    Output:\\
    \upshape
    (\string"relationship\string"\{tuple\_delimiter\}Alex\{tuple\_delimiter\}Taylor\{tuple\_delimiter\}Alex is affected by Taylor's authoritarian certainty and observes changes in Taylor's attitude towards the device.\{tuple\_delimiter\}7)\{record\_delimiter\}\\
    (\string"relationship\string"\{tuple\_delimiter\}Alex\{tuple\_delimiter\}Jordan\{tuple\_delimiter\}Alex and Jordan share a commitment to discovery, which contrasts with Cruz's vision.\{tuple\_delimiter\}6)\{record\_delimiter\}\\
    (\string"relationship\string"\{tuple\_delimiter\}Taylor\{tuple\_delimiter\}Jordan\{tuple\_delimiter\}Taylor and Jordan interact directly regarding the device, leading to a moment of mutual respect and an uneasy truce.\{tuple\_delimiter\}8)\{record\_delimiter\}\\
    (\string"relationship\string"\{tuple\_delimiter\}Jordan\{tuple\_delimiter\}Cruz\{tuple\_delimiter\}Jordan's commitment to discovery is in rebellion against Cruz's vision of control and order.\{tuple\_delimiter\}5)\{record\_delimiter\}\\
    (\string"relationship\string"\{tuple\_delimiter\}Taylor\{tuple\_delimiter\}The Device\{tuple\_delimiter\}Taylor shows reverence towards the device, indicating its importance and potential impact.\{tuple\_delimiter\}9)\{completion\_delimiter\}\\[6pt]
    
    \ttfamily
    \#\#\#\#\#\#\#\#\#\#\#\#\#\#\#\#\#\#\#\#\#\#\\
    \ldots\\
    \#\#\#\#\#\#\#\#\#\#\#\#\#\#\#\#\#\#\#\#\#\#\\
    \ldots\\
    \#\#\#\#\#\#\#\#\#\#\#\#\#\#\#\#\#\#\#\#\#\#\\
    
    -Real Data-\\
    \#\#\#\#\#\#\#\#\#\#\#\#\#\#\#\#\#\#\#\#\#\#\\
    Entities: \{entities\}\\
    Text: \{input\_text\}\\
    \#\#\#\#\#\#\#\#\#\#\#\#\#\#\#\#\#\#\#\#\#\#\\
    Output:
    \end{flushleft}
    
    \end{AIbox}
    \caption{The prompt template designed to extract relations from entities and text chunks.}
    \label{prompt_re}
    \end{figure*}

    \begin{figure*}[!h]
    \centering
    \scriptsize
    \begin{AIbox}{{Summary Entity and Relation Extraction}}
    
    \begin{flushleft}
    \ttfamily
    -Goal-\\
    You are tasked with analyzing a set of entity descriptions and a given list of meta attributes. Your goal is to summarize at least one attribute entity for the entity set in the given entity descriptions. And the summarized attribute entity must match the type of at least one meta attribute in the given meta attribute list (e.g., if a meta attribute is "company", the attribute entity could be "Amazon" or "Meta", which is a kind of meta attribute "company"). And it should be directly relevant to the entities described in the entity description set. The relationship between the entity set and the generated attribute entity should be clear and logical.\\[4pt]
    
    -Steps-\\
    1. Identify at least one attribute entity for the given entity description list. For each attribute entity, extract the following information:\\
    - entity\_name: Name of the entity, capitalized\\
    - entity\_type: One of the following types: [\{meta\_attribute\_list\}], normal\_entity means that doesn't belong to any other types.\\
    - entity\_description: Comprehensive description of the entity's attributes and activities\\
    Format each entity as (\string"entity\string"\{tuple\_delimiter\}\textless entity\_name\textgreater\{tuple\_delimiter\}\textless entity\_type\textgreater\{tuple\_delimiter\}\textless entity\_description\textgreater)\\[4pt]
    
    2. From each given entity, identify all pairs of (source\_entity, target\_entity) that are *clearly related* to the summary entities identified in step 1. And there should be no relations between the summary entities.\\
    For each pair of related entities, extract the following information:\\
    - source\_entity: name of the source entity, as given in entity list\\
    - target\_entity: name of the target entity, as identified in step 1\\
    - relationship\_description: explanation as to why you think the source entity and the target entity are related to each other\\
    - relationship\_strength: a numeric score indicating strength of the relationship between the source entity and target entity\\
    Format each relationship as (\string"relationship\string"\{tuple\_delimiter\}\textless source\_entity\textgreater\{tuple\_delimiter\}\textless target\_entity\textgreater\{tuple\_delimiter\}\textless \\ relationship\_description\textgreater\{tuple\_delimiter\}\textless relationship\_strength\textgreater)\\[4pt]
    
    3. Return output in English as a single list of all the entities and relationships identified in steps 1 and 2. Use **\{record\_delimiter\}** as the list delimiter.\\[4pt]
    
    4. When finished, output \{completion\_delimiter\}\\[6pt]
    
    \#\#\#\#\#\#\#\#\#\#\#\#\#\#\#\#\#\#\#\#\#\#\\
    -Examples-\\
    \#\#\#\#\#\#\#\#\#\#\#\#\#\#\#\#\#\#\#\#\#\#\\
    Example 1:\\
    
    Input:\\
    Meta summary entity list: ["company", "location"]\\
    Entity description list: [("Instagram", "Instagram is a software developed by Meta, which captures and shares the world's moments. Follow friends and family to see what they're up to, and discover accounts from all over the world that are sharing things you love."), ("Facebook", "Facebook is a social networking platform launched in 2004 that allows users to connect, share updates, and engage with communities. Owned by Meta, it is one of the largest social media platforms globally, offering tools for communication, business, and advertising."), ("WhatsApp", "WhatsApp Messenger: A messaging app of Meta for simple, reliable, and secure communication. Connect with friends and family, send messages, make voice and video calls, share media, and stay in touch with loved ones, no matter where they are")]\\
    \#\#\#\#\#\#\#\\
    Output:\\
    \upshape
    ("entity"{tuple\_delimiter}"Meta"{tuple\_delimiter}"company"{tuple\_delimiter}"Meta, formerly known as Facebook, Inc., is an American multinational technology conglomerate. It is known for its various online social media services.")\{record\_delimiter\}\\
    ("relationship"{tuple\_delimiter}"Instagram"{tuple\_delimiter}"Meta"{tuple\_delimiter}"Instagram is a software developed by Meta."{tuple\_delimiter}8.5)\{record\_delimiter\}\\
    ("relationship"{tuple\_delimiter}"Facebook"{tuple\_delimiter}"Meta"{tuple\_delimiter}"Facebook is owned by Meta."{tuple\_delimiter}9.0)\{record\_delimiter\}\\
    ("relationship"{tuple\_delimiter}"WhatsApp"{tuple\_delimiter}"Meta"{tuple\_delimiter}"WhatsApp Messenger is a messaging app of Meta."{tuple\_delimiter}8.0)\{record\_delimiter\}\\
    \#\#\#\#\#\#\#\#\#\#\#\#\#\#\#\#\#\#\#\#\#\#\\
    Example 2:\\
    \ldots\\
    \#\#\#\#\#\#\#\#\#\#\#\#\#\#\#\#\#\#\#\#\#\#\\
    Example 3:\\
    \ldots\\
    \#\#\#\#\#\#\#\#\#\#\#\#\#\#\#\#\#\#\#\#\#\#\\
    
    -Real Data-\\
    \#\#\#\#\#\#\#\#\#\#\#\#\#\#\#\#\#\#\#\#\#\#\\
    Input:\\
    Meta summary entity list: \{meta\_attribute\_list\}\\
    Entity description list: \{entity\_description\_list\}\\
    \#\#\#\#\#\#\#\\
    Output:
    \end{flushleft}
    
    \end{AIbox}
    \caption{The prompt template designed to generate summary entities and the corresponding relations.}
    \label{prompt_ae}
    \end{figure*}

\subsection{Prompt Templates for Entity Extraction}
As shown in Figure~\ref{prompt_ee}, we used that prompt template to extract entities from text chunks. We also give three examples to guide the LLM to extract entities with higher accuracy.

\subsection{Prompt Templates for Relation Extraction}
As shown in Figure~\ref{prompt_re}, we extract relations from the entities extracted earlier and the corresponding text chunks. Then we can get the triples in the basic knowledge graph, which is also the 0-th layer of the hierarchical knowledge graph.

\subsection{Prompt Templates for Entity Summarization}
As shown in Figure~\ref{prompt_ae}, we generate summary entities in each layer of the hierarchical knowledge graph. We will not only let the LLM generate the summary entities from the previous layer, but also let it generate the relations between the entities of these two layers. These relations will clarify the reasons for summarizing these entities.

\subsection{Prompt Templates for RAG Evaluation}
In terms of the prompt templates we use to conduct evaluations, we utilize the same prompt design as that in LightRAG. The prompt will let the LLM generate both evaluation results and the reasons in JSON format to ensure clarity and accuracy.

\end{document}